\documentclass[conference]{IEEEtran}
\IEEEoverridecommandlockouts
\usepackage{cite}
\usepackage{amsmath,amssymb,amsfonts}
\usepackage{algorithmic}
\usepackage{graphicx}
\usepackage{textcomp}
\usepackage{xcolor}

\usepackage{algorithm}
\usepackage{algorithmic}
\usepackage{booktabs}

\def\BibTeX{{\rm B\kern-.05em{\sc i\kern-.025em b}\kern-.08em
    T\kern-.1667em\lower.7ex\hbox{E}\kern-.125emX}}
\begin{document}

\title{Sample-Efficient Multi-Agent Reinforcement Learning with Demonstrations for Flocking Control\\

}

\author{\IEEEauthorblockN{Yunbo Qiu, Yuzhu Zhan, Yue Jin, Jian Wang, Xudong Zhang}
\IEEEauthorblockA{
Department of Electronic Engineering, Tsinghua University, Beijing, China \\
\{qyb18, zhanyz16, jiny16\}@mails.tsinghua.edu.cn, \{jian-wang, zhangxd\}@tsinghua.edu.cn}
}

\IEEEpubid{\copyright~2022 IEEE}

\maketitle

\begin{abstract}
Flocking control is a significant problem in multi-agent systems such as multi-agent unmanned aerial vehicles and multi-agent autonomous underwater vehicles, which enhances the cooperativity and safety of agents. In contrast to traditional methods, multi-agent reinforcement learning (MARL)  solves the problem of flocking control more flexibly.
However, methods based on MARL suffer from sample inefficiency, since they require a huge number of experiences to be collected from interactions between agents and the environment.
We propose a novel method Pretraining with Demonstrations for MARL (PwD-MARL), which can utilize non-expert demonstrations collected in advance with traditional methods to pretrain agents. During the process of pretraining, agents learn policies from demonstrations by MARL and behavior cloning simultaneously, and are prevented from overfitting demonstrations. By pretraining with non-expert demonstrations, PwD-MARL improves sample efficiency in the process of online MARL with a warm start.
Experiments show that PwD-MARL improves sample efficiency and policy performance in the problem of flocking control, even with bad or few demonstrations.

\end{abstract}


\section{Introduction}
Flocking control is an important problem in multi-agent systems, applied in many areas such as multi-agent unmanned aerial vehicles, and multi-agent autonomous underwater vehicles. In the problem of flocking control, agents in a multi-agent system are required to navigate to a target area without collisions or moving far from each other.
Collision avoidance, velocity matching, and flock centering are three rules for flocking control \cite{reynolds1987flocks}.

Various traditional methods have been designed to solve the problem of flocking control. For example, artificial potential field methods \cite{sun2019hybrid,sharma2009flocking} build attractive and repulsive fields to guide agents fulfill corresponding subgoals.
Leader-follower methods \cite{yu2010distributed, shao2020leader} select an agent as a leader and instruct other agents to follow the leader.
Virtual structure methods \cite{fan2013bipartite, yu2013flocking} require agents to maintain a rigid formation. However, these traditional methods demand sophisticated policies designed for agents to comply with various task constraints.

With the development of deep learning, methods based on reinforcement learning (RL) have been proposed for the problem of flocking control. With the excellent feature representation capability of deep neural networks and precise instructions of rewards from interactions between agents and the environment, RL methods can generate more flexible and more powerful policies. In the problem of flocking control, single-agent-based methods \cite{ wang2018deep, yan2020fixed,xu2018multi} are applied to each agent independently, where for each single agent other agents are considered as a part of the environment. However, these methods suffer from a non-stationary environment. Therefore, MARL-based methods \cite{zhu2020multi, zhao2020research} are further proposed, which take mutual influences of agents' varying policies into account during learning.

However, when these methods are used in the problem of flocking control, a considerable number of interaction samples between agents and the environment are required, due to the complexity of this problem. The process of collecting experience samples is expensive, and thus it is crucial to improve sample efficiency of RL algorithms. 

To improve sample efficiency of RL-based methods, some methods leverage the idea of learning from demonstrations.
These methods mainly include behavior cloning \cite{pomerleau1988alvinn}, inverse reinforcement learning \cite{ziebart2008maximum}, and generative adversarial imitation learning \cite{ho2016generative}. They can also be extended to multi-agent cases \cite{le2017coordinated,natarajan2010multi,song2018multi}.
In addition, \cite{nair2018overcoming, goecks2020integrating,vinyals2019grandmaster} combine behavior cloning with RL. 
However, most of these methods demand a huge dataset with expert demonstrations. An exception is  \cite{nair2018overcoming}, which claims that a few non-expert demonstrations can also help its training. Nevertheless, in a complex environment such as flocking control, the performance of this method declines a lot, which is shown in Section \uppercase\expandafter{\romannumeral5}.

\IEEEpubidadjcol

We propose a novel sample-efficient MARL method, which utilizes demonstrations to pretrain agents. We name it Pretraining with Demonstrations for MARL (PwD-MARL).
Before agents interact with the environment, a few non-expert demonstrations are used to help agents pretrain their policies.
During the pretraining process, agents simultaneously imitate demonstrations and optimize their policies with an RL algorithm in an offline manner. Besides, agents are prevented from overfitting the previously collected demonstrations.
Owing to our pretraining method, agents can take the advantage of the demonstrations to achieve a warm start before interacting with the environment. It helps the agents learn preliminary knowledge of the task, and thus learn faster with fewer interaction samples. 
In addition, PwD-MARL doesn't set rigid limitations on the quality and quantity of demonstration data.  
We leverage PwD-MARL to solve the problem of flocking control. Experiments show that PwD-MARL helps agents learn faster than the pure online MARL algorithm \cite{lowe2017multi} and an algorithm incorporating RL and learning from demonstrations \cite{nair2018overcoming}. Experiments also verify that PwD-MARL can perform well when demonstrations are much worse in performance or fewer in quantity.

The main contributions of this paper are listed as follows:
\begin{itemize}
\item A novel algorithm PwD-MARL is proposed to improve sample efficiency of MARL by utilizing non-expert demonstrations for pretraining.
\item Extensive experiments show that PwD-MARL can solve the flocking control problem with both higher sample efficiency and better performance, even with bad or few demonstrations.
\end{itemize}

\section{Background}

\subsection{Markov Game}

We consider flocking control as a Markov Game in this paper. At each time step $t$, each agent $i$ receives an observation $o_i$ from the environment, and chooses an action $a_i$ according to its policy $\pi_i$ to interact with the environment. The joint observation of all the agents is denoted as $\boldsymbol{o}$, and the joint action is denoted as $\boldsymbol{a}$.
The environment gives each agent a reward $r_i$ and each agent receives a new observation  $o_i^{\prime}$.
The goal is to maximize the cumulative rewards of each agent, i.e. $\Sigma_{t=0}^{+\infty} \gamma^{t} r_{i,t}$, where $\gamma$ is a discount factor. 

\subsection{Multi-Agent Reinforcement Learning Algorithm}

There are three commonly applied components in MARL algorithms to assist the training process: the actor-critic framework, replay buffers, and centralized training and decentralized execution (CTDE).

In the actor-critic framework, each agent has an ‘actor’ function and a ‘critic’ function. The ‘actor’ function, also known as policy function $\pi_{i}(o_i)$, generates an action $a_i$ according to local observation $o_i$, and then the agent implements the action $a_i$ to interact with the environment. The ‘critic’ function is generally an action-value function \cite{sutton2018reinforcement}, also known as Q-function, which estimates the expected return of a policy given a particular state and action. 

A replay buffer is used to store transition samples, where a sample is represented as a tuple $<\boldsymbol{o},\boldsymbol{a}, r, \boldsymbol{o^{\prime}}>$. Agents can reuse these experiences to learn. During training, a minibatch of experiences is randomly sampled from the replay buffer in each iteration.

CTDE is commonly used in MARL. During the training process, observations of other agents are also provided for a certain agent. It enables the agent to better estimate its Q-function with global information. When it comes to the execution process, observations of other agents are no longer available. Each agent generates its action based on its local observation.

MADDPG \cite{lowe2017multi} is a representative MARL algorithm. For agent $i$, the loss of its policy function is:
\begin{equation}
\begin{aligned}
L_{i, actor}&=\mathbb{E}_{\boldsymbol{o},\boldsymbol{a}\sim \mathcal{D}}[-Q_i(\boldsymbol{o},a_1,...,a_i,...,a_n)|_{a_i=\pi_i(o_i)}],\\
\end{aligned}
\label{equ-actor-on}
\end{equation}
where $\mathcal{D}$ denotes the replay buffer. 
The loss of its Q-function is: 
\begin{equation}
\begin{aligned}
L_{i, critic}&=\mathbb{E}_{\boldsymbol{o},\boldsymbol{a},r_i,\boldsymbol{o^{\prime}}\sim \mathcal{D}}[(Q_i (\boldsymbol{o},a_1,...,a_n)-y_{i})^2],\\
\end{aligned}
\label{equ-critic-on}
\end{equation}
where $y_{i}$ is the target defined as: 
\begin{equation}
\begin{aligned}
y_{i}&=r_i+\gamma Q'_i (\boldsymbol{o}',a'_1,...,a'_n)|_{a'_j=\pi'_j(o'_j)},\\
\end{aligned}
\label{equ-y-on}
\end{equation}
where $Q'_i$ and $\pi'_j$ are target functions of Q-function and policy function, respectively. These target functions are softly updated to help stabilize the training process.

\section{Problem Formulation}

\subsection{Observations and Actions}

\begin{figure}[t]
\begin{center}
\centerline{\includegraphics[width=0.35\textwidth]{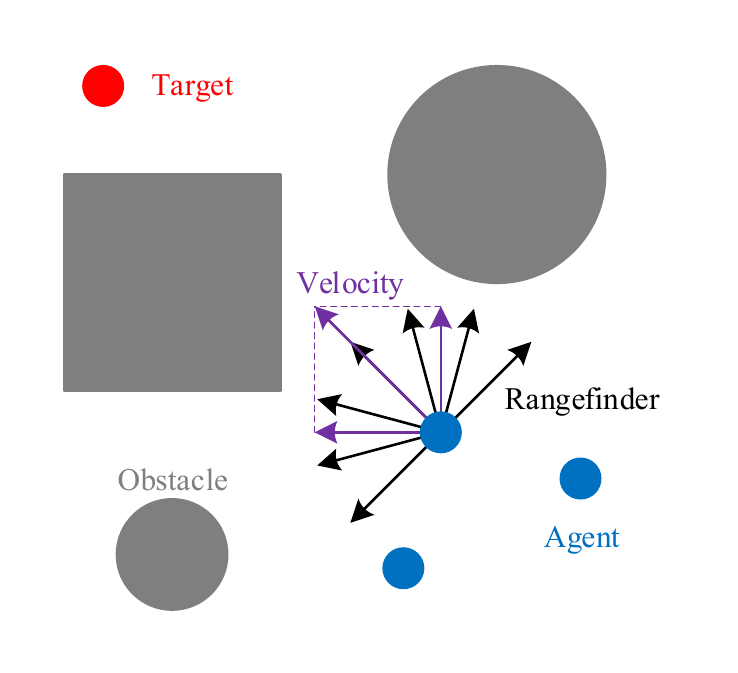}}
\end{center}
\caption{Observations of agents.}
\label{fig1}
\end{figure}

As shown in Fig.~\ref{fig1}, observations of an agent consist of three parts: relative positions of the target and other agents, the detected distances between obstacles and the agent (7 rangefinders are evenly deployed in the front of the agent every 30 degrees), and the horizontal and vertical speed of itself. Actions of an agent consist of the magnitude and angle of the force applied to the agent.

\subsection{Reward Scheme}

For agent $i$ at time $t$, the agent receives a reward $r_{i,t}$ from the environment after the agent takes an action. $r_{i,t}$ has 6 compositions corresponding to various subgoals of the problem of flocking control:
\begin{equation}
\begin{aligned}
r_{i,t}=&\rho_{nav}\cdot{r_{nav}}+\rho_{flock}\cdot{r_{flock}}+\rho_{col}\cdot{r_{col}}\\
&+\rho_{cross}\cdot{r_{cross}}+\rho_{time}\cdot{r_{time}}+\rho_{stab}\cdot{r_{stab}},\\
\end{aligned}
\end{equation}
where $\rho_{nav}$, $\rho_{flock}$, $\rho_{col}$, $\rho_{cross}$, $\rho_{time}$, and $\rho_{stab}$ are coefficients of various reward terms.

Specifically, $r_{nav}$ is the navigation reward to instruct the agent to move towards the target, which is defined as:
\begin{equation}
\begin{aligned}
r_{nav}&=d_{tar} (o_{i,t-1})-d_{tar} (o_{i,t}),\\
\end{aligned}
\end{equation}
where $d_{tar} (o_{i,t})$ denotes the distance between agent $i$ and the target at time $t$, as a part of observations mentioned before.

$r_{flock}$ is the flocking reward to instruct the agent not to move farther from the center of the agents than $th_{f}$: 
\begin{equation}
\begin{aligned}
r_{flock}&=relu(d_f (o_{i,t-1})-th_{f})-relu(d_f (o_{i,t})-th_{f}),\\
\end{aligned}
\end{equation}
where $d_f (o_{i,t})$ denotes the distance between agent $i$ and the center of the flock at time $t$, and $relu(x)=max(0,x)$ is the linear rectification function. 

$r_{col}$ and $r_{cross}$ are the collision reward and crossing reward, respectively, defined as:
\begin{equation}
r_{col}=
\left\{
\begin{aligned}
&(d_{obs} (o_{i,t})-th_{col})^3,&&{\text{if }}d_{obs} (o_{i,t})<th_{col},\\
&0,&&{\text{otherwise}},\\
\end{aligned}
\right.
\end{equation}
\begin{equation}
r_{cross}=
\left\{
\begin{aligned}
&(d_{ag} (o_{i,t},j)-th_{cross})^3,&&{\text{if }}d_{ag} (o_{i,t},j)<th_{cross},\\
&0,&&{\text{otherwise}},\\
\end{aligned}
\right.
\end{equation}
where $d_{obs} (o_{i,t})$ and $d_{ag} (o_{i,t},j)$ denote the distance sensed by rangefinders between agent $i$ and the nearest obstacle, and the distance between agent $i$ and another agent $j$ at time $t$, respectively. $th_{col}$ and  $th_{cross}$ are distance thresholds to determine whether to penalize the agent.

$r_{time}$ and $r_{stab}$ are the time reward and stability reward to encourage the agent to complete the flocking control task as soon as possible and as stable as possible, which are defined as:
\begin{equation}
\begin{aligned}
r_{time}&=-1,\\
\end{aligned}
\end{equation}
\begin{equation}
\begin{aligned}
r_{stab}&=- F,\\
\end{aligned}
\end{equation}
where $F$ is the magnitude of the force applied to the agent.

\section{Algorithm}

\begin{algorithm} [t]
	\caption{PwD-MARL} 
	\label{PwD-MARL} 
	\begin{algorithmic}
	    \REQUIRE a buffer with demonstrations $\mathcal{P}$
	    \FOR {step $= 1 \text{ to } S_{pretrain}$ }
	    \FOR {agent $i = 1 \text{ to } n$ }
	    \STATE Randomly sample a mini-batch $M$ from $\mathcal{P}$
	    \STATE Update actor by minimizing (\ref{equ-actor-pretrain})
        \STATE Update critic by minimizing (\ref{equ-critic-pretrain})
        \ENDFOR
	    \STATE Update target networks for each agent $i$: 
	    $\theta '_i \leftarrow \tau \theta _i + (1-\tau) \theta '_i$
	    \ENDFOR
	    \FOR {episode $= 1 \text{ to } E$ }
	    \STATE Initialize a random process $\mathcal{N}$ for action exploration
        \STATE Receive initial observation $\boldsymbol{o}$
        \FOR{$t$ = 1 to $T_{episode}$}
        \STATE For each agent $i$, select action $a_i = \pi_{\theta_i}(o_i) + \mathcal{N}_t$ w.r.t. the current policy and exploration
        \STATE Execute actions $\boldsymbol{a} = (a_1, . . . , a_n )$ and observe reward $r$ and new observation $\boldsymbol{o}'$
        \STATE Store $(\boldsymbol{o}, \boldsymbol{a}, \boldsymbol{r}, \boldsymbol{o}')$ in replay buffer $\mathcal{D}$
        \STATE $\boldsymbol{o} \leftarrow \boldsymbol{o}'$
	    \FOR {agent $i = 1 \text{ to } n$ }
	    \STATE Randomly sample a mini-batch $M$ from $\mathcal{D}$
	    \STATE Update actor by minimizing (\ref{equ-actor-on})
        \STATE Update critic by minimizing (\ref{equ-critic-on})
	    \ENDFOR
	    \STATE Update target networks for each agent $i$: 
	    $\theta '_i \leftarrow \tau \theta _i + (1-\tau) \theta '_i$
	    \ENDFOR
	    \ENDFOR
	\end{algorithmic} 
\end{algorithm}

Traditional MARL methods that learn with the actor-critic framework require a huge number of samples to gradually optimize Q-functions and policy functions. The interactions between agents and the environment are usually expensive. Therefore, it's critical to improve sample efficiency of MARL.

In this paper, we propose a novel algorithm, Pretraining with Demonstrations for MARL (PwD-MARL), to improve sample efficiency, and thus speed up the training process and promote policy performance.
Specifically, we consider using demonstrations preserved in a buffer $\mathcal{P}$ to pretrain Q-functions and policy functions of agents, which results in a warm start for the following online training and thus improves the sample efficiency and accelerates learning. 
PwD-MARL can be built on MARL algorithms that use the actor-critic framework. In this paper, we present our algorithm based on a typical MARL algorithm MADDPG \cite{lowe2017multi}.

The format of demonstrations is the same as that of experiences stored in the replay buffer during the online training process. Therefore, during pretraining, the demonstrations including $<\boldsymbol{o},\boldsymbol{a}, r, \boldsymbol{o^{\prime}}>$ can help optimize the Q-functions in the same way as online RL. However, since demonstrations are limited in quantity and their performance may not be expert, we prevent Q-functions from overfitting these demonstrations. Otherwise, during online training, it will take a lot of time to rectify the overfitting of the pretrained Q-functions, which increases the training time. Specifically, the loss for the Q-function of agent $i$ during the pretraining process is designed as a combination of RL loss and overfitting loss:
\begin{equation}
\begin{aligned}
L_{i, critic}^{pretrain}=&\mathbb{E}_{\boldsymbol{o},\boldsymbol{a},r_i,\boldsymbol{o^{\prime}}\sim \mathcal{P}}[(Q_i (\boldsymbol{o},a_1,...,a_n)-y_{i})^2] \\
& + \alpha_{critic} \cdot \left | \left |  Qparam_{i}  \right | \right |_{2},\\
\end{aligned}
\label{equ-critic-pretrain}
\end{equation}
where $y_{i}$ is the target value defined as (\ref{equ-y-on}), $\left | \left |  Qparam_{i}  \right | \right |_{2}$ is the $l_2$ norm of the neural network parameters of agent $i$'s Q-function, and $\alpha_{critic}$ is a hyperparameter.
Note that the second term in (\ref{equ-critic-pretrain}) is not involved in the loss function of online training.

As for the policy functions, during the pretraining process, we expect policy functions to generate actions similar to the actions in demonstrations in the same states. 
On the one hand, in contrast to a randomly initialized policy, the policy of demonstrations is better in most cases, although it may not be expert. Therefore, behaving like demonstrations instructs policy functions to achieve better performance. On the other hand, without the constraint that policy functions generate actions similar to demonstrations, the distribution of agents' actions under a certain state will deviate from that of demonstrations. The deviation of distribution will cause inaccurate estimations of Q-functions, and further harm the performance of policy functions. 



The loss for the policy functions during the pretraining process is a combination of the RL loss term and a behavior cloning loss \cite{pomerleau1988alvinn} term:
\begin{equation}
\begin{aligned}
L_{i,actor}^{pretrain}&=L_{i,actor}^{pretrain}+\beta_{i}\cdot L_{i,bc}^{pretrain},\\
\end{aligned}
\label{equ-actor-pretrain}
\end{equation}
where $\beta_{i}$ is a parameter to decide the mix ratio of RL loss term and the behavior cloning loss term. $L^{pretrain}_{i, bc}$ is the behavior cloning loss designed as: 
\begin{equation}
\begin{aligned}
L_{i,bc}^{pretrain}&=\mathbb{E}_{\boldsymbol{o},\boldsymbol{a}\sim \mathcal{P}}[(\pi_i(o_i)-a_i)^2],\\
\end{aligned}
\label{bcloss}
\end{equation}
$L_{i,actor}^{pretrain}$ is similar to (\ref{equ-actor-on}) except that it extracts demonstrations from $\mathcal{P}$ instead of experiences from $\mathcal{D}$:
\begin{equation}
\begin{aligned}
L_{i, actor}^{pretrain}&=\mathbb{E}_{\boldsymbol{o},\boldsymbol{a}\sim \mathcal{P}}[-Q_i(\boldsymbol{o},a_1,...a_i,...,a_n)|_{a_i=\pi_i(o_i)}].\\
\end{aligned}
\label{equ-actor-on-P}
\end{equation}

To convert the behavior cloning loss term to the same scale as RL loss term, $\beta_{i}$ is automatically calculated similarly to \cite{fujimoto2021minimalist} as:
\begin{equation}
\begin{aligned}
\beta_{i}&=\frac{\alpha_{actor}}{M}\sum_{\boldsymbol{o},\boldsymbol{a} \sim \mathcal{P}}|Q_i(\boldsymbol{o},a_1,...,a_i,...,a_n)|_{a_i=\pi_i(o_i)}|,\\
\end{aligned}
\end{equation}
where $M$ is the size of the minibatch sampled from the demonstrations buffer, and $\alpha_{actor}$ is a hyperparameter.

Note that a considerable number of expert demonstrations are not necessary for PwD-MARL, since PwD-MARL only utilizes demonstrations during the pretraining process for a warm start, instead of continuously relying on demonstrations as in \cite{nair2018overcoming}. If a fixed collection of demonstrations are continuously used for training, the limitation of quantity will result in a severe overfitting problem, and the poor performance of demonstrations will restrain agents from learning better. In contrast, our way of utilizing demonstrations causes that PwD-MARL has loose restrictions on the quantity and quality of collected demonstrations.

The whole algorithm is presented in Algorithm \ref{PwD-MARL}.

\section{Experiments}

\subsection{Experimental Settings}

\begin{figure}[t]
\begin{center}
\centerline{\includegraphics[width=0.35\textwidth]{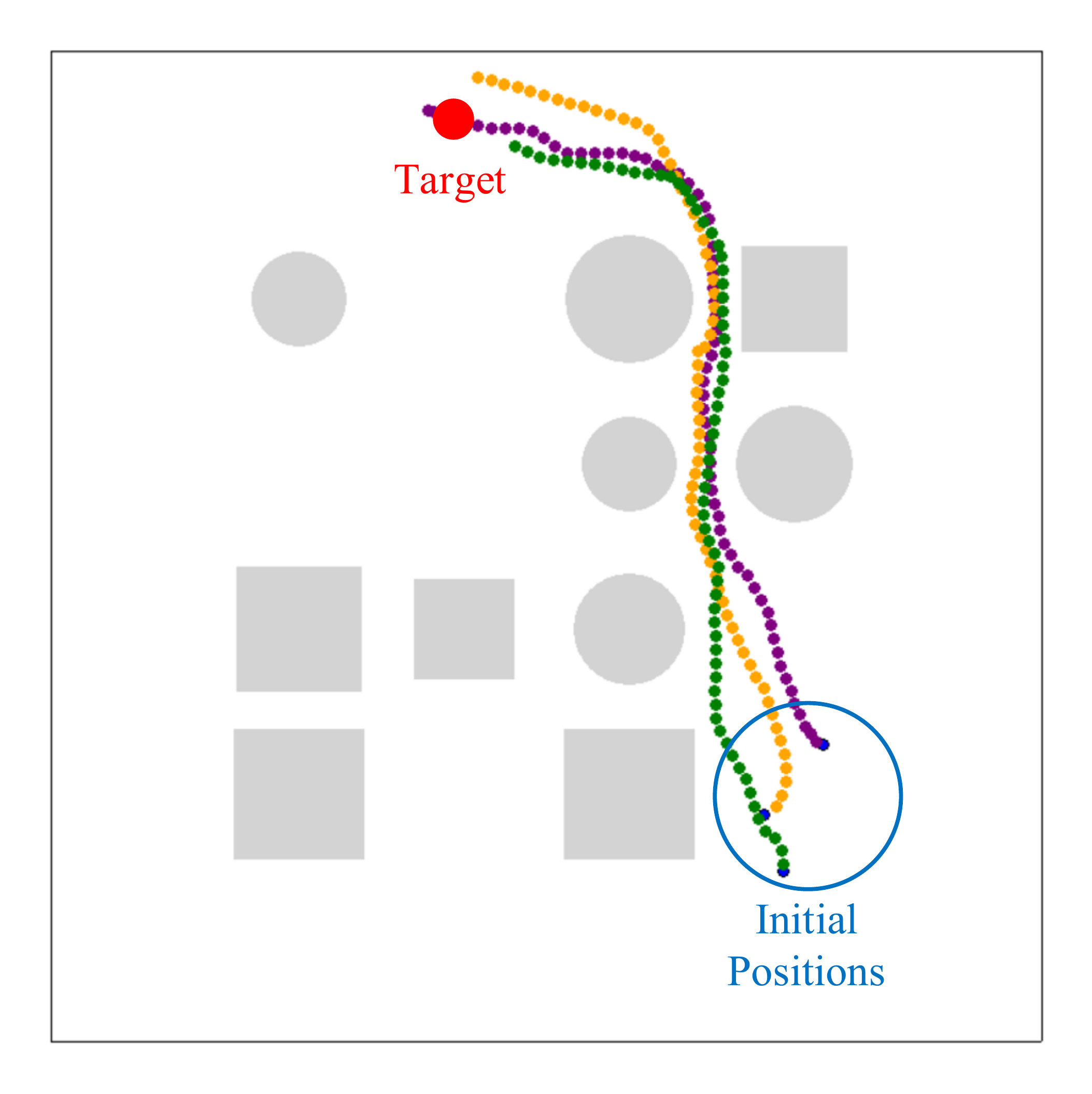}}
\end{center}
\caption{The environment and trajectories of an episode. The target is red and enlarged. Initial positions of agents are blue and circled. Trajectories of three agents are green, purple, and orange, respectively. Obstacles are grey.}
\label{fig2}
\end{figure}

An environment of flocking control is simulated to validate PwD-MARL. The map of the environment is square with a side length of $L~unit$. At the beginning of each episode, $n$ round agents are located in the middle area of the map, close to each other. $m$ round obstacles and other $m$ square obstacles are also located in the middle area of the map. The target is a circle area located near the edge of the map. Locations of the target, agents, and obstacles are randomly initialized in every episode. An example of the environment is shown in Fig.~\ref{fig2}.
An episode is successful only when the agents are all closer to the target than $d_{arrive}$ within $T_{episode}$ time steps without collisions. The acceleration and speed of the agents are capped.

In our experiment, the diameters of agents and the target are all 0.4 $unit$. The diameters or the side lengths of obstacles are uniformly randomized from 3 to 5 $unit$. The magnitude of acceleration and speed of the agents can not exceed 0.5 $unit$.

Neural networks of policy functions and Q-functions are all fully connected networks. The networks have 3 hidden layers with 64 units and a $tanh$ activation function in each layer. Adam optimizer is used for optimization, with a learning rate of 0.001.

Demonstrations are generated by an artificial potential field algorithm \cite{park2001obstacle} adapted to the environment of flocking control. We used 3000 episodes of demonstrations in our experiments, whose average success rate in the flocking control task is 0.803.
The capacity of the replay buffer $\mathcal{D}$ is 300000.
Results of all the algorithms are run with 3 different seeds.

Detailed values of other hyperparameters of environments and algorithms mentioned above in the paper are listed in Table \ref{tab0}.

The experiments are conducted by Python and TensorFlow. The environments are generated by Tkinter.


\begin{table}[t]
\caption{Hyperparameters of Environments and Algorithms}
\begin{center}
\begin{tabular}{cccc}
\toprule
\textbf{Hyperparameter} & \textbf{Value}  & \textbf{Hyperparameter} & \textbf{Value}\\ \midrule
     $L$          &   36       &     $\rho_{stab}$      &    $\frac{1}{L}$  \\
        $unit$       &    20  &   $th_{f}$         &    1.5 $unit$  \\
         $n$      &    3   &  $th_{col}$          &   1 $unit$ \\
          $m$      &    5  & $th_{cross}$         &  1.5 $unit$  \\
     $d_{arrive}$          &   3 $unit$   & $\gamma$    & 0.95 \\
     $T_{episode}$        &  100   &    $\tau$     &    0.0004 \\
         $\rho_{nav}$      &    $\frac{0.25}{L}$ & $\alpha_{critic}$  &  0.00002\\
        $\rho_{flock}$      &    $\frac{0.5}{L}$  & $\alpha_{actor}$   &    2.5\\
        $\rho_{col}$      &    $\frac{80}{L}$  & $S_{pretrain}$ & 300000\\
         $\rho_{cross}$      &    $\frac{40}{L}$  & $M$  &    32 \\
        $\rho_{time}$     &   $\frac{1}{L}$  &  $E$  &    200000 \\ \bottomrule

\end{tabular}
\label{tab0}
\end{center}
\end{table}

\subsection{Main Experiments}

In the problem of flocking control, we compare our algorithm PwD-MARL with three baseline algorithms: MADDPG, SVL-MARL, and MARLwD. MADDPG \cite{lowe2017multi} is an online RL algorithm, where we build our algorithm PwD-MARL. 
SVL-MARL is built according to \cite{vinyals2019grandmaster}, which utilizes supervised learning to learn from demonstrations during the process of pretraining.
MARLwD is an algorithm that simultaneously learns from demonstrations and online RL. We design MARLwD based on \cite{nair2018overcoming} for MARL problems. The algorithm in \cite{nair2018overcoming} doesn't require that demonstrations are expert, unlike \cite{le2017coordinated,natarajan2010multi, goecks2020integrating}. 

\begin{figure}[t]
\begin{center}
\centerline{\includegraphics[width=0.5\textwidth]{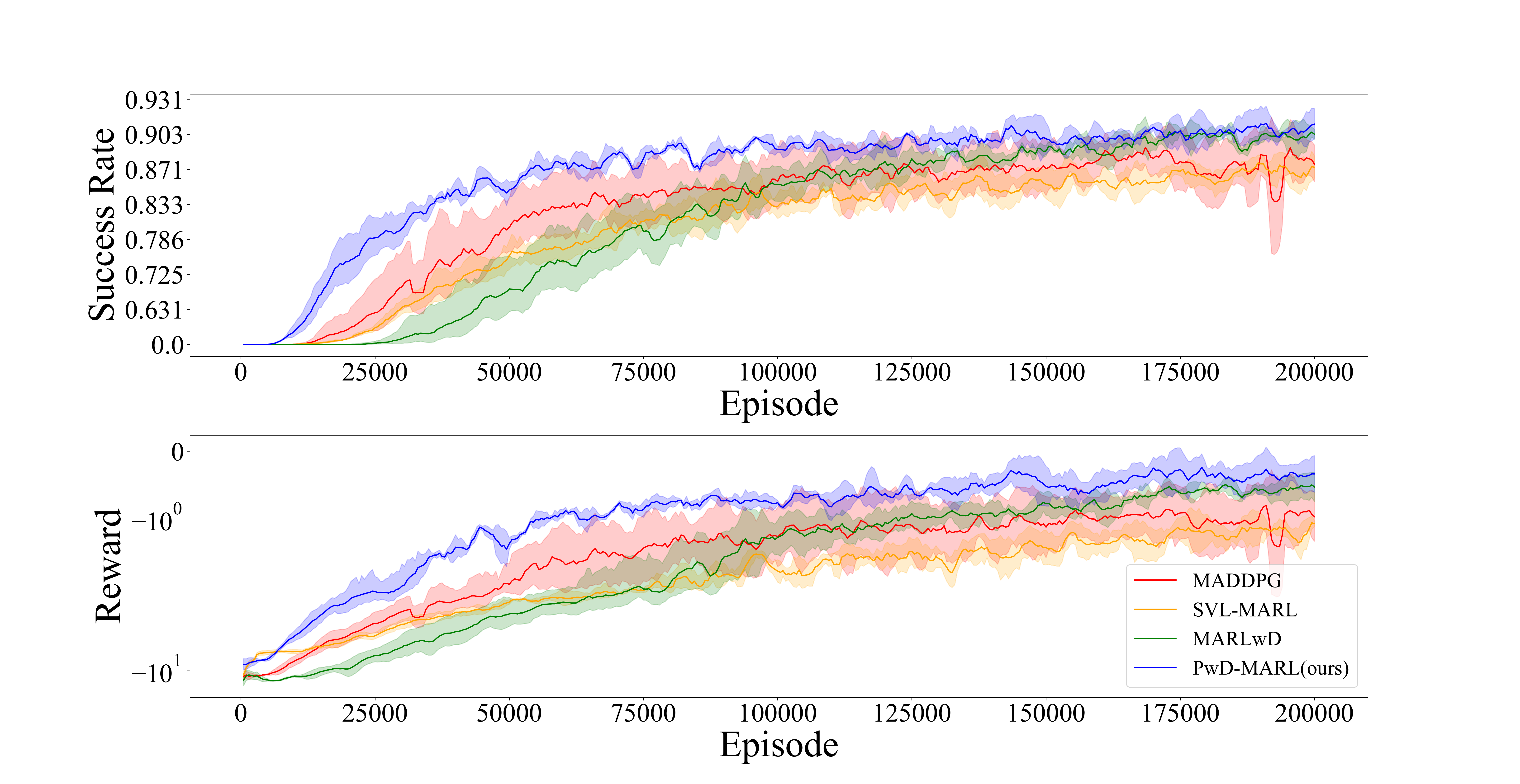}}
\end{center}
\caption{Convergence curves of success rate and reward of main experiments.}
\label{fig3}
\end{figure}

Success rate and reward are two main metrics that we focus on, since they represent the primary performance from the perspective of the task and RL, respectively.
Convergence curves of success rate and reward are shown in Fig.~\ref{fig3}, presented in the fifth root scale and the symmetric log scale, respectively. 
We can see that our algorithm PwD-MARL converges much faster than MADDPG, SVL-MARL and MARLwD, which shows the ability to improve sample efficiency of PwD-MARL.
On the one hand, PwD-MARL achieves better performance in success rate and reward than three baseline algorithms with the same number of episodes. On the other hand, PwD-MARL requires fewer episodes to achieve the same performance than three baseline algorithms.
We can also see from Fig.~\ref{fig3} that PwD-MARL surpasses MARLwD in performance, and both of them surpass MADDPG and SVL-MARL by a lot. It shows that non-expert demonstrations can help agents learn better policies in MARLwD and PwD-MARL, and our algorithm PwD-MARL utilizes demonstrations better than MARLwD.

We also test the midterm and final performance of these four algorithms over 2500 episodes. Detailed results are listed in Table \ref{tab2} and Table \ref{tab1}, respectively. 
The final performance is evaluated after 200000 training episodes, and the midterm performance is evaluated after 100000 training episodes. They can demonstrate comprehensive results with different limitations of sample numbers.
Apart from success rate and average reward of episodes, they also list the average distance between agents and the flock center, average time steps of episodes, and average magnitude of force applied to agents.
These three metrics are supplementary metrics that we can use to evaluate the comprehensive performance of algorithms.

As shown in Table \ref{tab2} and Table \ref{tab1}, with the same number of training episodes, PwD-MARL achieves a higher success rate and a higher reward, in comparison with MADDPG, SVL-MARL, and MARLwD.
As for flock distance, PwD-MARL can keep agents closer to each other according to both the midterm performance and the final performance.
In terms of time steps, PwD-MARL is superior to MADDPG and MARLwD, and only inferior to SVL-MARL. However, we should notice that success rate of SVL-MARL is much worse than the other three algorithms, which means that agents trained by SVL-MARL are more likely to collide and terminate the task in advance.
Therefore, PwD-MARL can complete the task with fewer time steps without harm to success rate.
When it comes to force, we can see that PwD-MARL requires less force to complete the task than three baseline algorithms, except that MADDPG requires slightly less force when agents are trained after 100000 episodes.
In summary, PwD-MARL outperforms the baseline algorithms in terms of main metrics, and generally improves performance in three supplementary metrics.

\begin{table*}[t]
\caption{Statistical Results After 100000 Training Episodes}
\begin{center}
\begin{tabular}{cccccccc}
\toprule
\textbf{Algorithm}                                    & \textbf{Success Rate} & \textbf{Reward}  & \textbf{Flock Distance} & \textbf{Time Steps} & \textbf{Force} \\ \midrule
\textbf{PwD-MARL (ours)}           & 0.890                 & -0.757                   & 1.846                   & 47.071              & 43.876         \\
\textbf{MADDPG}                    & 0.858                 & -1.269              & 1.948                   & 49.271              & 43.785         \\
\textbf{SVL-MARL}                    & 0.830                 & -1.774              & 1.909                   & 45.869             & 56.475        \\
\textbf{MARLwD}                    & 0.855                 & -1.274             & 2.096                   & 54.216              & 49.665         \\ \midrule
\textbf{no BC (ablation)}          & 0.882                 & -0.867                   & 1.891                   & 47.316              & 41.461         \\
\textbf{no RL (ablation)}          & 0.841                 & -1.455                 & 1.916                   & 46.866              & 49.253         \\
\textbf{no overfitting (ablation)} & 0.867                 & -1.180                 & 1.890                   & 46.950              & 47.934          \\ \bottomrule
\end{tabular}
\label{tab2}
\end{center}
\end{table*}

\begin{table*}[t]
\caption{Statistical Results After 200000 Training Episodes}
\begin{center}
\begin{tabular}{cccccccc}
\toprule
\textbf{Algorithm}                                    & \textbf{Success Rate} & \textbf{Reward}  & \textbf{Flock Distance} & \textbf{Time Steps} & \textbf{Force} \\ \midrule
\textbf{PwD-MARL (ours)}           & 0.912                 & -0.331                   & 1.821                   & 47.624              & 41.979         \\
\textbf{MADDPG}                    & 0.876                 & -0.966              & 1.933                   & 48.339              & 44.009         \\
\textbf{SVL-MARL}                    & 0.872                 & -1.070              & 1.895                   & 46.409              & 49.942        \\
\textbf{MARLwD}                    & 0.903                 & -0.525             & 1.972                   & 50.847              & 47.089         \\ \midrule
\textbf{no BC (ablation)}          & 0.901                 & -0.535                   & 1.861                   & 47.141              & 42.038         \\
\textbf{no RL (ablation)}          & 0.874                 & -0.957                 & 1.889                   & 47.156              & 45.793         \\
\textbf{no overfitting (ablation)} & 0.872                 & -0.931                 & 1.894                   & 47.246              & 46.14          \\ \bottomrule
\end{tabular}
\label{tab1}
\end{center}
\end{table*}

\subsection{Ablation Experiments}

\begin{figure}[t]
\begin{center}
\centerline{\includegraphics[width=0.5\textwidth]{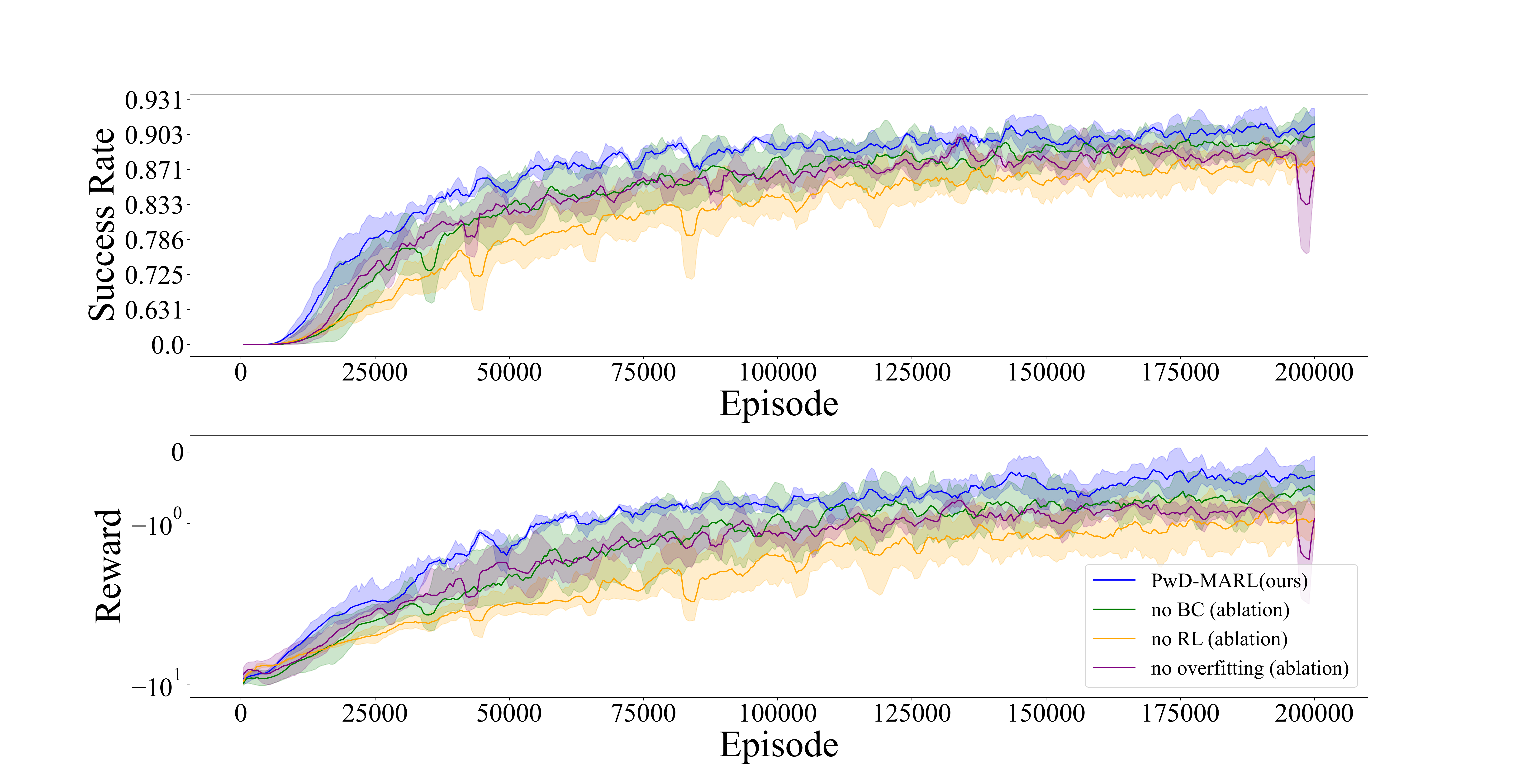}}
\end{center}
\caption{Convergence curves of success rate and reward of ablation experiments.}
\label{fig4}
\end{figure}

To illustrate the effectiveness of all the parts of our proposed algorithm, three ablation algorithms of PwD-MARL are designed. During the pretraining process, behavior cloning loss term in policy functions, reinforcement learning loss term in policy functions, or overfitting loss term in Q-functions is removed in each of three ablation algorithms. We call them ‘no BC’, ‘no RL’, and ‘no overfitting’, respectively.

Convergence curves are shown in Fig.~\ref{fig4}. We can see that Pwd-MARL outperforms all the ablation algorithms both in sample efficiency and in the performance of success rate and reward. It validates that all the parts of PwD-MARL are necessary.

Detailed statistics of the midterm and final performance are also listed in Table \ref{tab2} and Table \ref{tab1} for further analyses on the impact on comprehensive metrics.
We can see that besides the performance in success rate and reward, all three ablation algorithms are inferior to PwD-MARL in terms of flock distance.
Time steps of PwD-MARL and its three ablation algorithms are nearly the same, due to the impact of success rate on time steps, as we analyze in main experiments.
PwD-MARL also requires less force than three ablation algorithms after 100000 and 200000 training episodes, except the ablation algorithm ‘no BC’ after 100000 training episodes.
Overall, PwD-MARL is superior to its three ablation algorithms in terms of main metrics, and generally superior to ablation algorithms in terms of supplementary metrics.

\subsection{Experiments with Worse or Fewer Demonstrations}

\begin{figure}[t]
\begin{center}
\centerline{\includegraphics[width=0.5\textwidth]{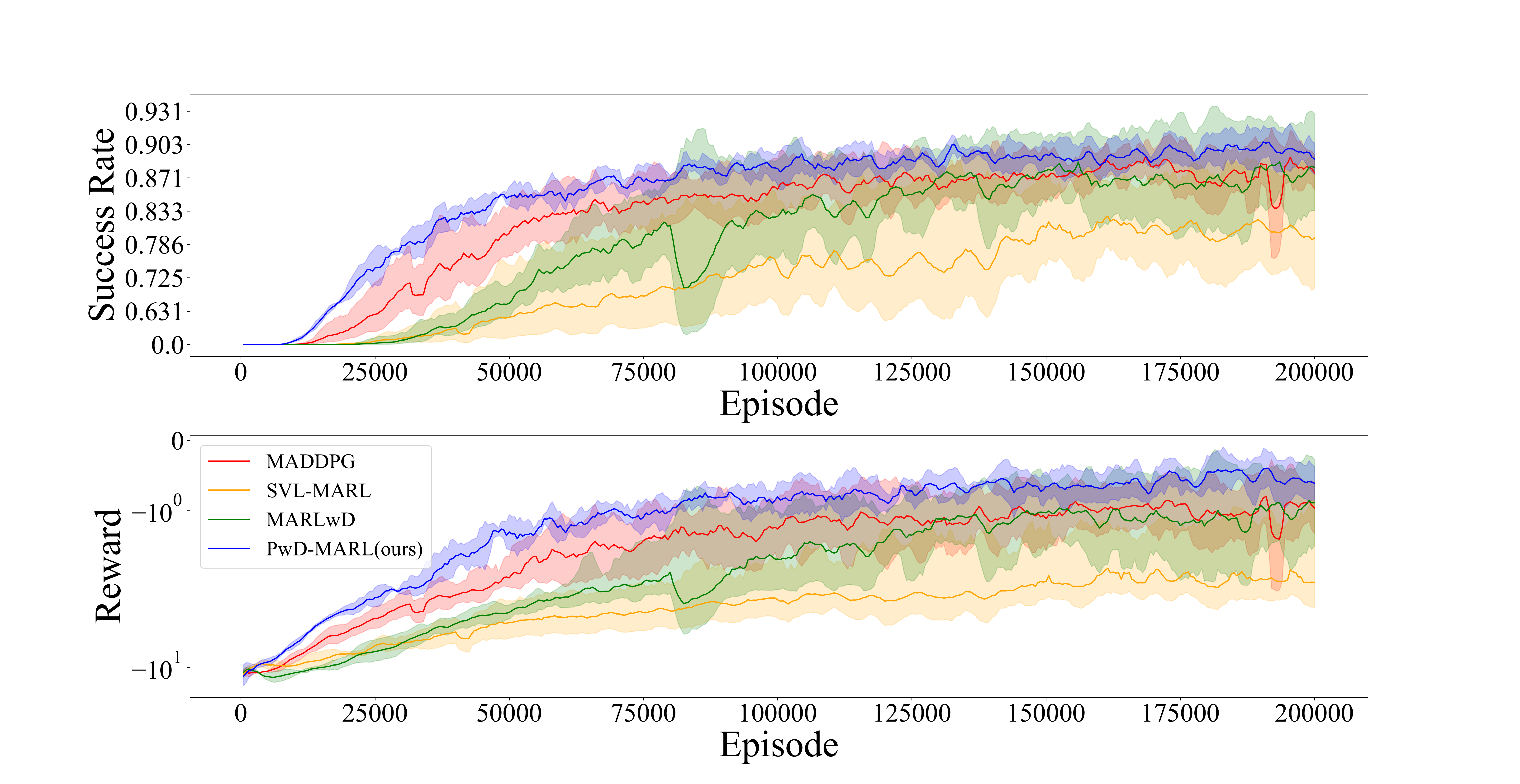}}
\end{center}
\caption{Convergence curves of success rate and reward of experiments with worse demonstrations.}
\label{fig5}
\end{figure}

\begin{figure}[t]
\begin{center}
\centerline{\includegraphics[width=0.5\textwidth]{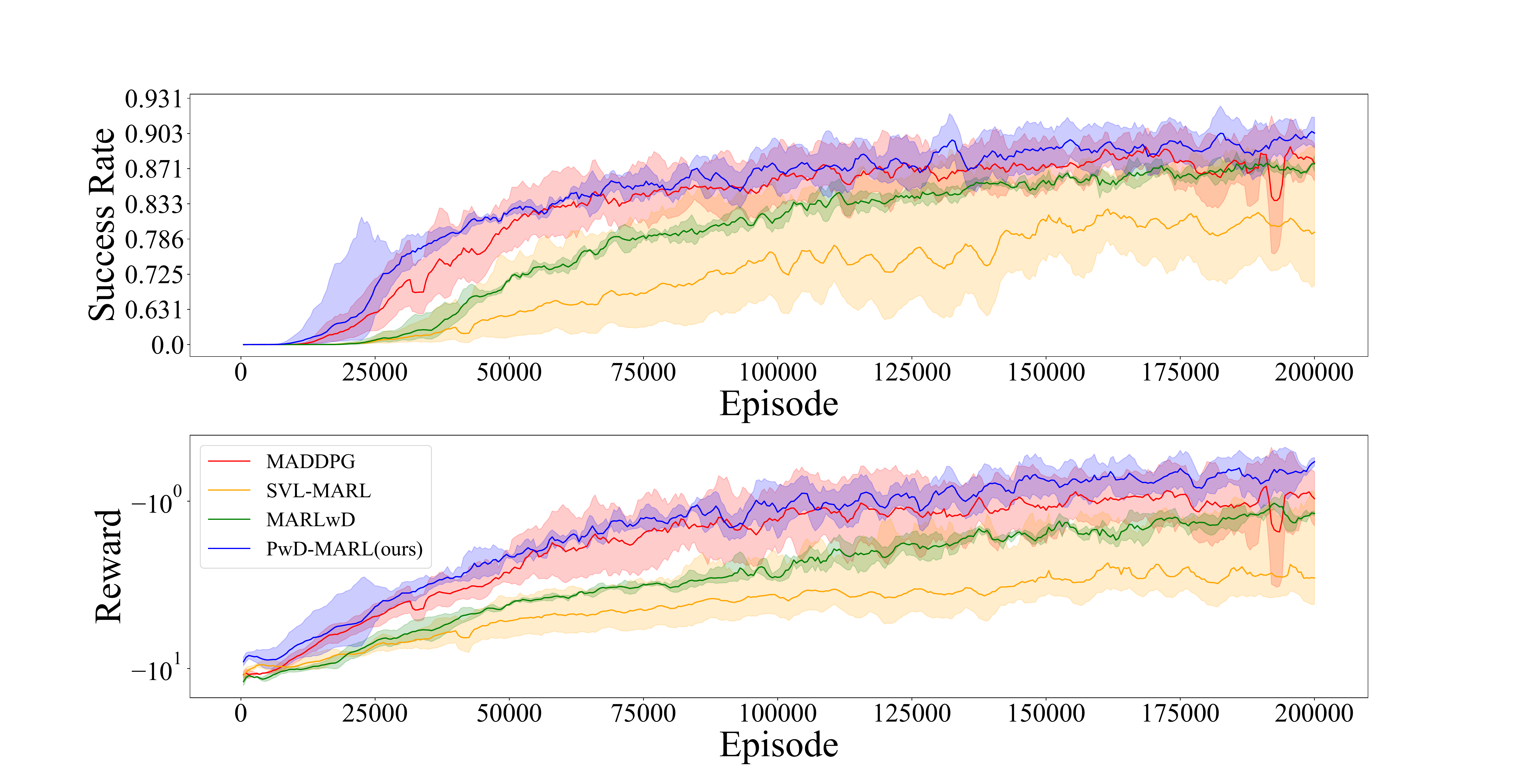}}
\end{center}
\caption{Convergence curves of success rate and reward of experiments with fewer demonstrations.}
\label{fig6}
\end{figure}

We conduct further studies to investigate the performance of our PwD-MARL algorithm with worse or fewer demonstrations.
For experiments with worse demonstrations, we use demonstrations whose success rate is 0.724 in the flocking control task, in contrast to a success rate of 0.803 in previous experiments.
For experiments with fewer demonstrations, we use 1500 episodes of demonstrations, in contrast to 3000 episodes in previous experiments.
Convergence curves are shown in Fig.~\ref{fig5} and Fig.~\ref{fig6}, respectively.

It can be seen from Fig.~\ref{fig5} and Fig.~\ref{fig6} that the performance of PwD-MARL in success rate and reward is better than three baseline algorithms with the same number of training episodes. Therefore, PwD-MARL can improve sample efficiency in comparison with MADDPG, even with worse or fewer demonstrations. 
PwD-MARL also improves final performance when demonstrations are worse or fewer.
In contrast, the performance of MARLwD and SVL-MARL is even worse than MADDPG, which shows that worse or fewer demonstrations even harm learning in MARLwD and SVL-MARL.
With further comparison of Fig.~\ref{fig3} and Fig.~\ref{fig5} as well as Fig.~\ref{fig6}, we can see that the quality and quantity of demonstrations only slightly hinder the performance of PwD-MARL, but largely reduce the performance of MARLwD and SVL-MARL.
These results show that PwD-MARL has loose restrictions on the quality and quantity of demonstrations, unlike MARLwD and SVL-MARL.

\section{Conclusions and Future Work}

In this paper, we propose a novel algorithm PwD-MARL and use PwD-MARL to solve the problem of flocking control. PwD-MARL utilizes non-expert demonstrations to pretrain agents' policy functions and Q-functions, which improves sample efficiency and policy performance of the algorithm. 
Experiments show that PwD-MARL can solve the problem of flocking control better than the baseline algorithms in success rate, sample efficiency, flock distance, time steps, and applied force. It's also shown that PwD-MARL has loose restrictions on the quality and quantity of demonstrations in contrast to other algorithms utilizing demonstrations.

Although we validate the effectiveness of PwD-MARL in the problem of flocking control in this paper, PwD-MARL is a general algorithm designed to improve the sample efficiency of MARL. In the near future, we expect to apply PwD-MARL to other applications.

\bibliographystyle{IEEEtran}
\bibliography{ref}

\end{document}